# Mesh Learning for Classifying Cognitive Processes

Mete Özay[a], Ilke Öztekin[b], Uygar Öztekin[c], Fatoş T. Yarman Vural[a]

[a]Department of Computer Engineering, Middle East Technical University, Ankara, Turkey.

[b]Department of Psychology, Koç University, Istanbul, Turkey.

[c]Google Inc. , California, USA.

## Abstract

A relatively recent advance in cognitive neuroscience has been multi-voxel pattern analysis (MVPA), which enables researchers to decode brain states and/or the type of information represented in the brain during a cognitive operation. MVPA methods utilize machine learning algorithms to distinguish among types of information or cognitive states represented in the brain, based on distributed patterns of neural activity. In the current investigation, we propose a new approach for representation of neural data for pattern analysis, namely a Mesh Learning Model. In this approach, at each time instant, a star mesh is formed around each voxel, such that the voxel corresponding to the center node is surrounded by its p-nearest neighbors. The arc weights of each mesh are estimated from the voxel intensity values by least squares method. The estimated arc weights of all the meshes, called Mesh Arc Descriptors (MADs), are then used to train a classifier, such as Neural Networks, k-Nearest Neighbor, Naïve Bayes and Support Vector Machines. The proposed Mesh Model was tested on neuroimaging data acquired via functional magnetic resonance imaging (fMRI) during a recognition memory experiment using categorized word lists, employing a previously established experimental paradigm (Öztekin & Badre, 2011). Results suggest that the proposed Mesh Learning approach can provide an effective algorithm for pattern analysis of brain activity during cognitive processing.

## 1. Introduction

One of the main obstacles in studying the human mind is the problem of representation. How is information represented in the brain, and how does the representation change depending on the type and nature of the information? This has long been an intriguing and challenging question in the history of science across many disciplines.

Recently, advances in pattern recognition and neuroscience have provided researchers access to neural data that reflects brain activity during deployment of cognitive processes, enabling a whole new range of complementary methods and approaches to study the human mind (Norman et al., 2006; Polyn et al., 2005; Wang et al., 2003; Haynes and Rees, 2006; Mitchell et al., 2004; Van De Ville and Lee, 2011). Multi-voxel pattern analysis (MVPA) has enabled researchers to infer the degree to which a type of information or a cognitive process is represented in the brain at a given time, based on distributed patterns of neural activation.

An important step for MVPA is to extract relevant features acquired from neuroimaging methods, such as functional magnetic resonance imaging (fMRI). The feature space enables representation of the type of information at a given stage during cognitive processing. Once the raw data is preprocessed, a common approach has been feeding the feature vector formed by the voxel intensity values to one of the well-known classifiers or clustering algorithms, such as Kernel Machines, Neural Networks, Bayesian classifiers or Ensemble Classifiers.

In order to find a powerful representation of the fMRI measurements several noise reduction and feature extraction approaches have been reported in the literature. For instance, a widely used approach for representation of fMRI data is to identify and select the active voxels for the condition of interests. A group of methods to select the informative voxels employs Independent Component Analysis (ICA) and Principal Component Analysis (PCA). Both approaches capture the spatial and temporal dependencies in fMRI measurements (Lautrup, Hansen, Law, Morch, Svarer & Strother, 1995; Rasmussen, Abrahamsen, Madsen & Hansen, 2012). Another approach is to identify active voxels for the experimental conditions against a control-baseline condition using the General Linear Model (GLM) (e.g. Pereira, Mitchell & Botvinick, 2009). This approach can further be improved via employing an information mapping technique called Searchlight (Kriegeskorte, Goebel, Bandettini (2006), which combines the signals from a voxel and its immediately adjacent neighbors to identify informative regions. Accordingly, Information Mapping can detect how well the multivariate signal in a local region can differentiate between experimental conditions.

In the current study, we suggest a new complementary approach, called Mesh Learning for pattern analysis of neuroimaging data. In this approach, neural activity corresponding to a particular type of information/cognitive state is represented as a collection of meshes. The arc weights of these meshes describe the interconnections among the voxels and can be estimated from patterns of activation acquired via functional Magnetic Resonance Imaging (fMRI). This approach is similar to Searchlight in that it considers the signal from a voxel and its neighbors. In Searchlight, informative voxels are found based on the combination of signals across all voxels in a specific region for appropriate feature selection. In the Mesh Learning model, however, the relationship between the neural activity of the neighboring voxels are modeled and directly fed to the classification algorithm. That is, rather than using this relationship for feature selection, the Mesh Model directly incorporates the information representing the relationship between the neural activation of the neighbor voxels into the classification algorithm.

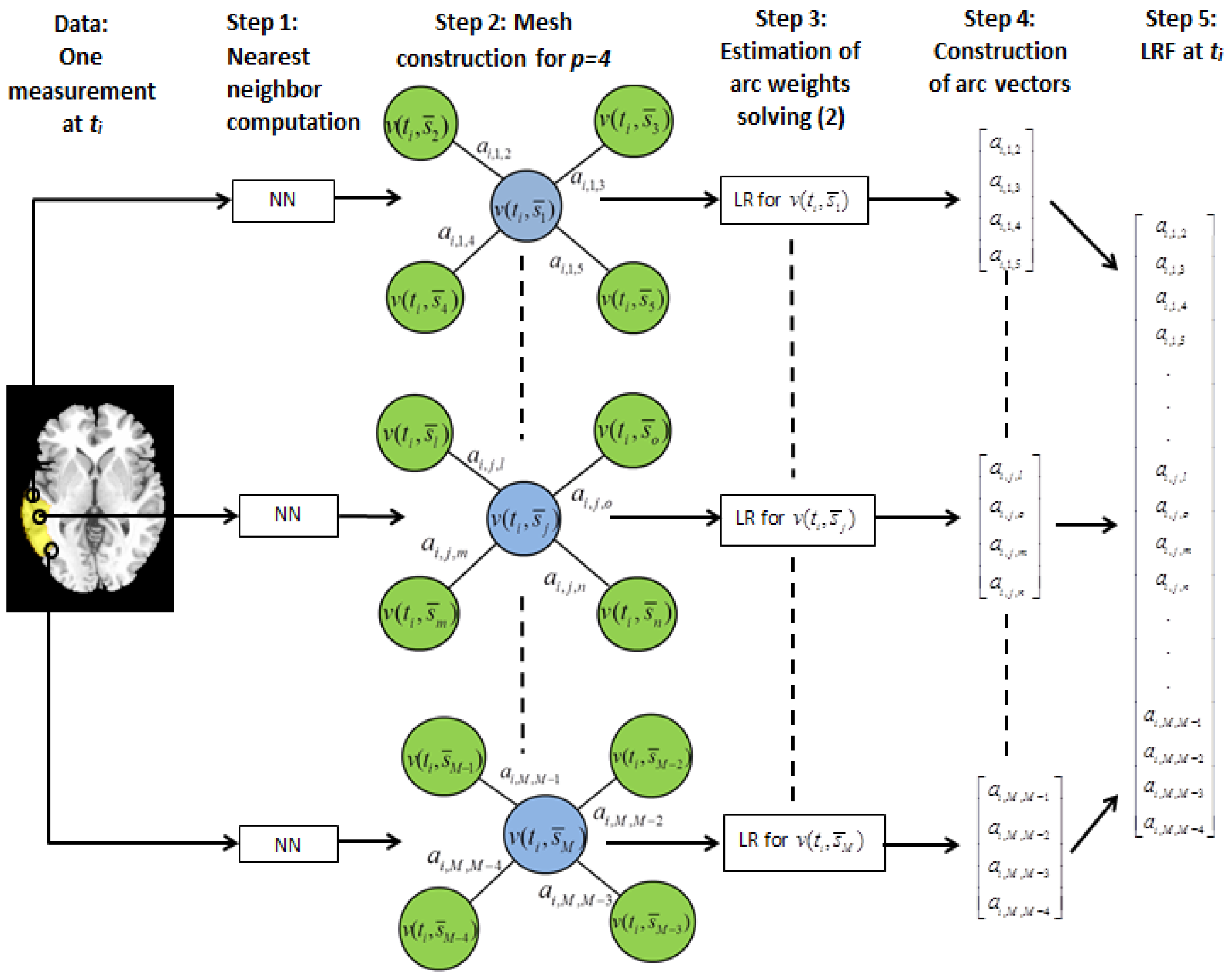

*Figure 1.* Illustration of the Mesh Learning approach. First step; we find the *p-nearest neighbor* of each voxel, using a spatial distance measure. Second step; we form a local mesh around each voxel, in a star topology where the surrounding voxels (green nodes) are the *p-nearest neighbors* of the center voxel (blue node). The mesh is defined for each voxel form the feature space of a classifier. Third step; the arc weights between the center voxel and its neighbors, in the mesh, are estimated by a linear regression model of the voxel intensity values, at a time instant (corresponding to the response of voxels to a predefined cognitive input). Fourth step; for each center voxel, a *p-dimensional* vector is formed with the entries of estimated arc weights. Fifth step; for each time instant, the arc weight vector for all the voxels are concatenated under a *Mxp* dimensional vector (MAD), where *M* is the number of active voxels.

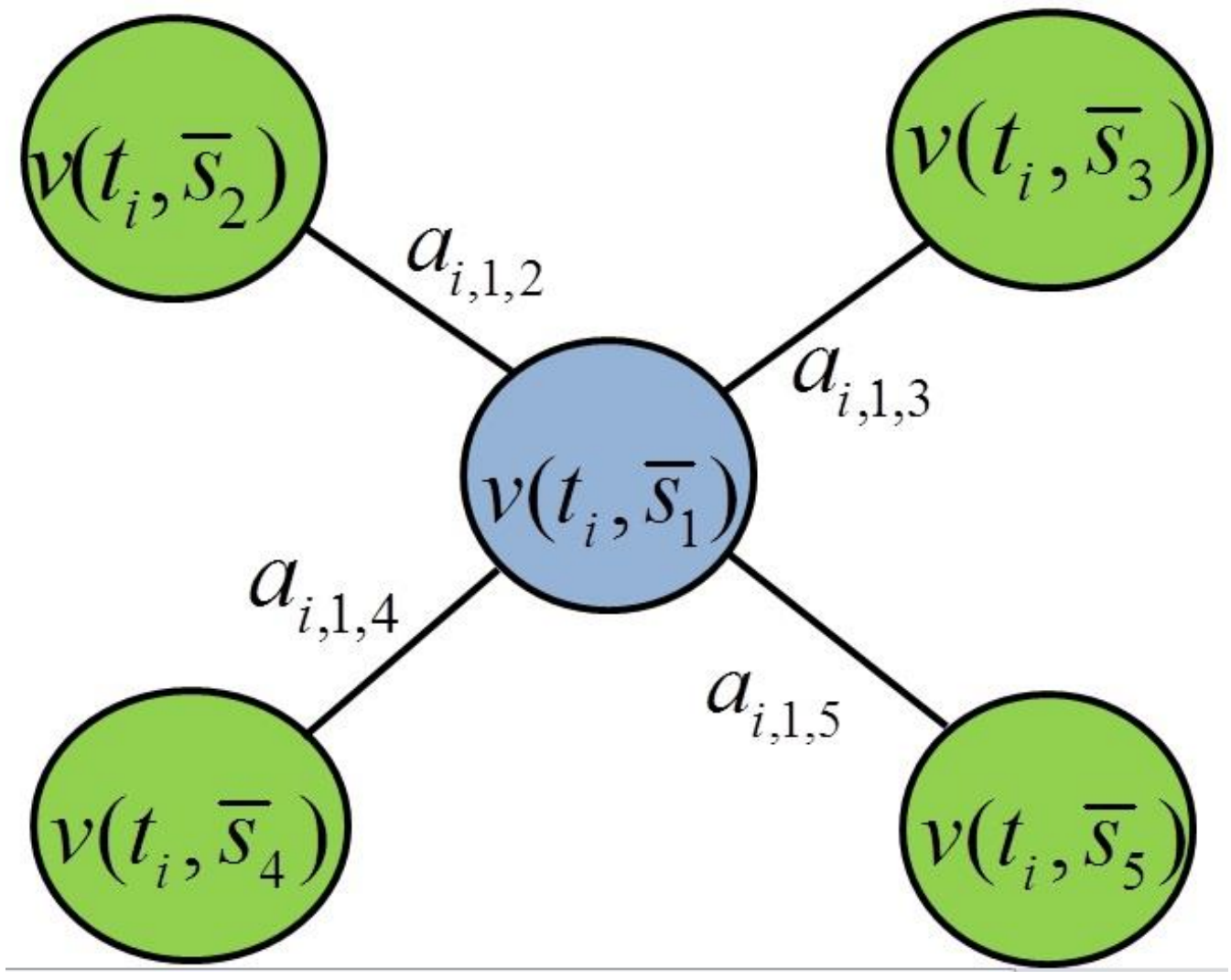

*Figure 2*. The Star Mesh, which represents the voxel intensity values $v(t_i, \bar{s}_k)$ at the center and its 4-nearest neighbors. Blue node represents the center voxel and the green nodes represent the surrounding voxels.

The steps of the Mesh Learning are shown in Figure 1. In the first step, the *p-nearest neighbors* of each voxel are found using a spatial distance measure. In the second step, a local mesh is formed around each voxel in star topology (see Figure 2), where the surrounding voxels (green nodes in Figure 2) are the p-nearest neighbors of the center voxel (blue node in Figure 2). In the third step, the arc weights between the center voxel intensities and its surroundings in the mesh are computed via least squares estimation. In the fourth step, for each center voxel, a *p-dimensional* vector is formed with the entries of estimated arc weights. Finally, the arc weight vectors computed from each mesh are concatenated under an *Nxp* dimensional vector, which we call the Mesh Arc Descriptor (MAD), where *N* is the number of active voxels, and *p* is the size of the mesh that defines the relationship of a voxel with its neighboring voxels. The MAD vectors, which represent the voxel interactions, can then be fed to widely used classifiers, such as Support Vector Machines, Neural Network, Naïve Bayes classifier and K-Nearest Neighbor classifiers.

Here we tested whether the proposed Mesh Learning approach can successfully predict the type of information represented in the brain during memory encoding and retrieval, using a previously established paradigm (Öztekin & McElree, 2007; Öztekin, Curtis & McElree; Öztekin & Badre, 2011), In a previous study (Öztekin & Badre, 2011), successful classification performance was obtained in predicting the semantic category of words during memory encoding and retrieval using a Neural Network classifier. In the current investigation, employing the proposed Mesh Model, we tested whether the success with which a classifier can predict the type of information represented in the brain during a cognitive operation could be improved.

The results implicate that the proposed representation yields higher discriminative power compared to PCA, ICA, GLM and Searchlight methods. Accordingly, we propose that a supervised learning system such as the Mesh Learning model suggested in this study could provide a useful tool to decode brain activation during cognitive processing.

## 2. Materials and Methods

### *2.1 Participants*

One right-handed female adult (age 22) participated in the experiment. The participant had normal or corrected-to-normal vision, and was screened for use of CNS affecting drugs, for psychiatric or neurological conditions, and for contraindications for MRI.

### *2.2 Design and Procedure*

Stimuli consisted of 21 instances of 10 semantic categories from the category norms of (Van Overschelde, Rawson and Dunlosky, 2004). The experiment consisted of eight 10-minute runs. Each run contained 30 experimental trials, in which participants studied a 5-item list, solved four math problems, and made a recognition judgment to a test word.

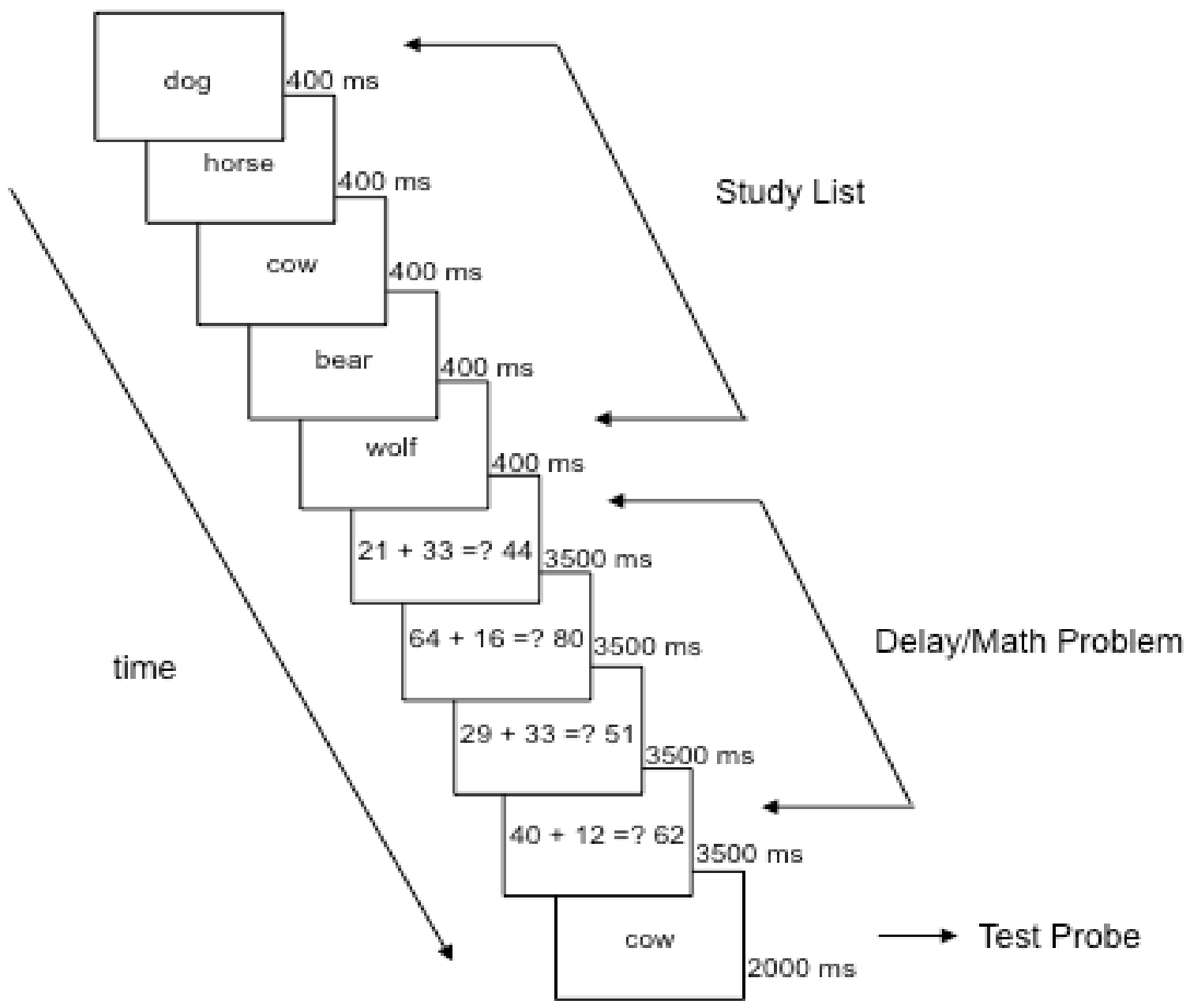

*Figure 3.* Illustration of a sample experimental trial sequence.

Figure 3 presents a sequence of events within a single experimental trial. Each trial proceeded with the presentation of a 5-word study list sequentially for 400 ms each. Following the $5^{th}$ word, participants solved four math problems consisting of addition or subtraction of two randomly selected 2-digit numbers. Participants indicated whether the solution presented next to the math problem was accurate by pressing either the middle or index finger on the button box. Participants had 3500 ms to respond to each math problem. Following the fourth math problem, participants were presented with a test word and indicated whether the word was a member of the current study list. Participants had 2000 ms to respond to the test probe. The inter-trial interval consisted of presentation of a fixation cross on the center of the screen for variable duration (ranging from 0 to 8000 ms).

Each trial proceeded with the presentation of a 5-word study list sequentially for 400 ms each. Following the 5$^{th}$ word, participants solved four math problems consisting of addition or subtraction of two randomly selected 2-digit numbers. Participants indicated whether the solution presented next to the math problem was accurate by pressing either the middle or index finger on the button box. Participants had 3500 ms to respond to each math problem. Following the fourth math problem, participants were presented with a test word and indicated whether the word was a member of the current study list. Participants had 2000 ms to respond to the test probe. The inter-trial interval consisted of presentation of a fixation cross on the center of the screen for variable duration (ranging from 0 to 8000 ms).

Critically, study lists and probes consisted of words that belonged to the same semantic category (e.g. animals) (Öztekin & McElree, 2007, 2010; Öztekin, Curtis & McElree, 2009; Öztekin and Badre, 2011). A total of ten semantic categories were used in the study. This way, we were able to employ supervised learning algorithms to classify the examples from one of the semantic categories of words the participant was encoding and retrieving from memory.

## *2.3 FMRI protocol, Image processing, and Data analysis*

Whole-brain imaging was performed on a Siemens 3T Magnetom Trio MRI system. Functional images were acquired using a gradient-echo echo-planar sequence, followed by high-resolution T1-weighted (MP-RAGE) anatomical images were collected for visualization. Head motion was restricted using firm padding that surrounded the head. Visual stimuli were projected onto a screen, and viewed through a mirror attached to a standard head coil.

Image processing and data analysis were performed using SPM5 (http://www.fil.ion.ucl.ac.uk/spm/). Following quality assurance procedures to assess outliers or artifacts in volume and slice-to-slice variance in the global signal, functional images were corrected for differences in slice acquisition timing by resampling all slices in time to match the first slice, followed by motion correction across all runs (using sinc interpolation). Functional data were then normalized based on MNI stereotaxic space using a 12-parameter affine transformation along with a nonlinear transformation using cosine basis functions. Images were resampled into 2-mm cubic voxels and then spatially smoothed with an 8-mm FWHM isotropic Gaussian kernel. Next, the functional data were detrended to account for baseline shifts across runs, and for scanner drift across the entire session for pattern analysis.

Previous research has implicated the lateral temporal cortex in storage and retrieval of semantic information (Damasio, 1990; Badre & Wagner, 2002; Thompson-Schill, 2003). Accordingly, classifiers were trained to distinguish the examples from ten semantic categories used in the experiment based on the distributed activation in lateral temporal cortex. The lateral temporal cortex ROI was structurally defined, and consisted of the left middle and left inferior temporal gyri. A set of regressors, which assigned each functional scan to a particular classification condition (i.e. the specific semantic category) assuming a lagged hemodynamic response function were determined. Consistent with previous research (e.g., Polyn et al., 2005, Öztekin & Badre, 2011), onsets were shifted forward by three points to account for the hemodynamic response lag, and a standard leave-one-out (N-minus-one) run-by-run cross validation approach was used for the pattern analysis. Classification success was determined by the proportion of correct category classifications across the test trials.

### *2.4 Descriptive analysis and motivation for the Mesh Model*

An initial descriptive analysis on the neuroimaging data suggested that intensity measures across the ten classes might not be sufficient in providing the desired discriminative power among the ten semantic categories. This fact can be observed from Figure 4, where at each time $t_i$, the participant is either encoding or retrieving items belonging to one of the ten semantic categories. The class conditional densities for the semantic categories have high overlaps suggesting that the statistical learning machines trained by the feature vectors consisting of the voxel intensity values might not have the desired discriminative power for classifying the semantic categories. On the other hand, a slight variation is observed on the intensity values along the *x*, *y* and *z* axis.

Figure 5 shows 2-dimensional distributions of voxel intensity values, for a time instant t, corresponding to a sample from category *animals*. Note that the intensity values smoothly vary in the neighboring voxels, in x, y and z directions. This observation motivated us to model the relationships among the voxel intensity values at each class. Since the variations are quite smooth, the relationship can be assumed to be locally linear and can be modeled by a linear regression equation in a pre-defined neighborhood system. In the following section, we modeled the spatial dependencies among the voxels.

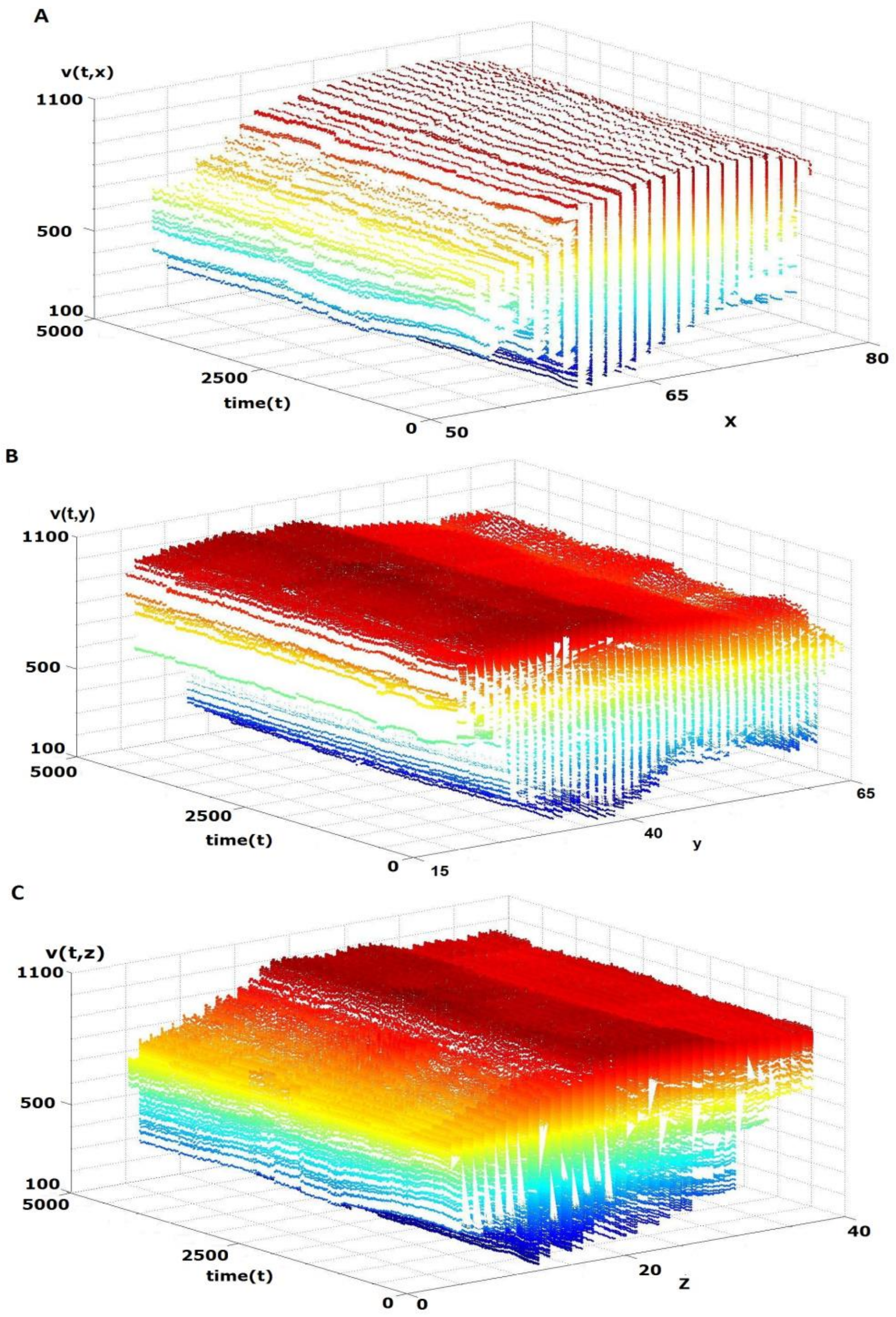

*Figure 4.* Spatio-temporal distributions of the voxel intensity values for each spatial dimension. Voxels are distributed on x-axis in (A), y-axis in (B), and z-axis in (C). The time axis indicates the fMRI intensity values with 10 different classes. The fMRI intensity values are quantized into 32 levels, each indicated by a color (low to high color wavelengths correspond to low to high of fMRI intensities) for visualization purposes.

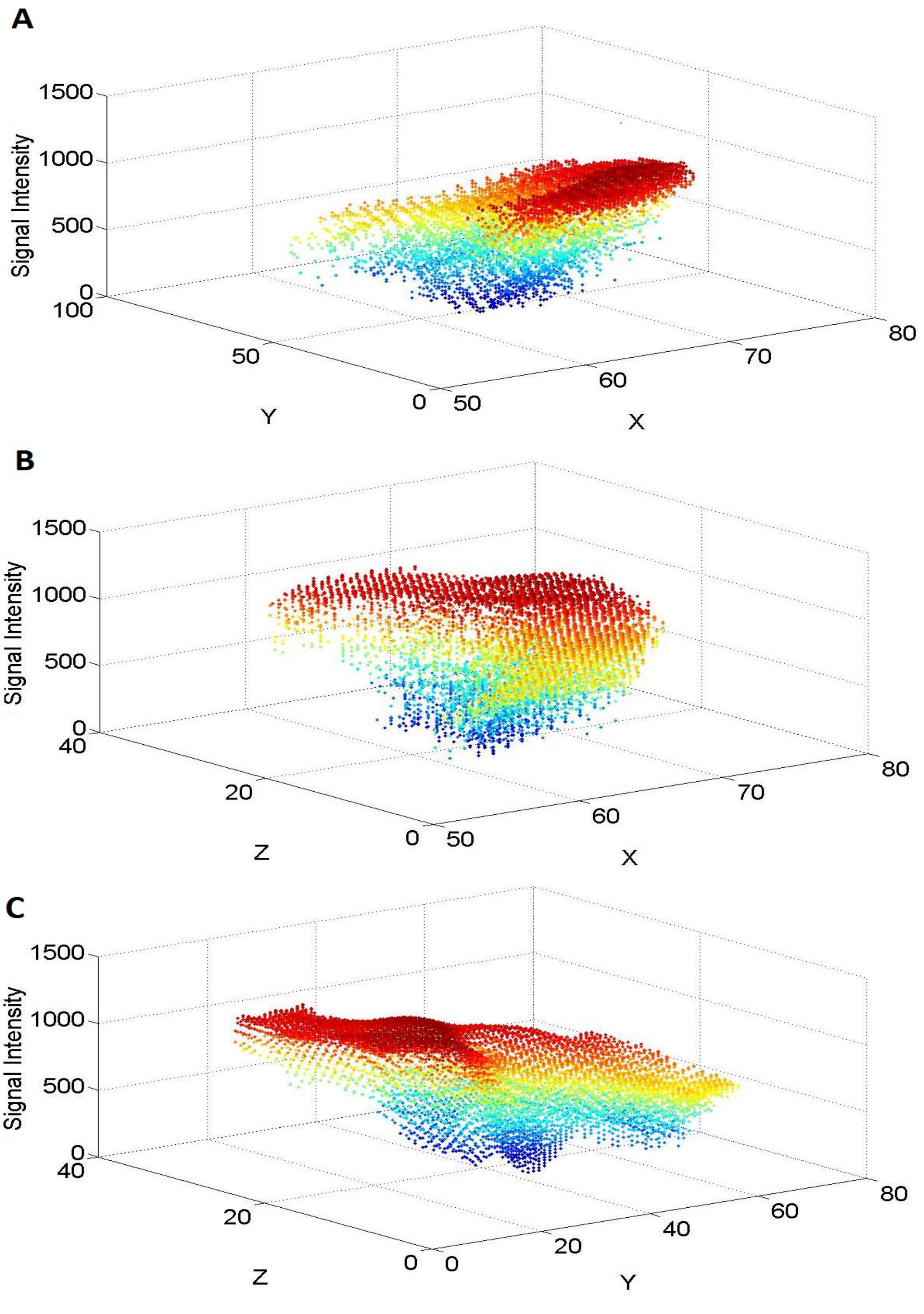

*Figure 5.* Spatial distribution of the voxel intensity values for a time instant t, corresponding to category *animal*. (A) Distribution in *x-y* directions, (B) Distribution in *x-z* directions, (C) Distribution in *z-y* directions. Intensity values are quantized into 32 levels to visualize the voxel intensity distribution using color.

## 3. Mesh Learning Model

In this study, the multiple voxel intensity values measured at a time instant denote an example which belongs to one of the ten semantic categories. The data set $D=\{v(t_i,\overline{s}_j): j=1,...,N, i=1,...,M\}$ contains the intensity values, $v(t_i,\overline{s}_j)$, which are measured at each voxel at location $\overline{s}_j$ at time instant $t_i$. The voxels are distributed in the brain in three dimensions, hence $\overline{s}_j=(x_j,y_j,z_j)$ is a three dimensional vector.

We start by defining a *p-neighborhood* system in the data set *D*. We define the *p-neighbor* of a voxel with intensity value $v(t_i,\overline{s}_j)$ as the *p-nearest neighbor*, at time $t_i$. Mathematically speaking, the nearest neighbor of a voxel $v(t_i,\overline{s}_j)$ is defined as,

$$\eta_1\left[v(t_i,\overline{s}_j)\right]=\{v(t_i,\overline{s}_j): \|\overline{s}_j-\overline{s}_k\|\leq\|\overline{s}_j-\overline{s}_l\|,\ \forall\ v(t_i,\overline{s}_j)\in D\},$$

where $\|.\|$ indicates the $l_2$ norm operation. Then, the *p-neighborhood* of a voxel $v(t_i,\overline{s}_j)$ is generated from the *(p-1)-neighborhood* iteratively, selecting the nearest neighbor of that voxel from $\eta_{p-1}\left[v(t_i,\overline{s}_j)\right]^c$, where *c* indicates the complement set of $\eta_{p-1}$. *p-nearest neighbors* of the voxel $v(t_i,\overline{s}_j)$ are obtained by adding the voxels in $\eta_{p-1}\left[v(t_i,\overline{s}_j)\right]$ to the nearest neighbor of $\eta_p$, as follows,

$$\eta_p\left[v(t_i,\overline{s}_j)\right]=\{v(t_i,\overline{s}_j)\cup\eta_{p-1}\left[v(t_i,\overline{s}_j)\right]: \|\overline{s}_j-\overline{s}_k\|\leq\|\overline{s}_j-\overline{s}_l\|, \forall v(t_i,\overline{s}_j)\in\eta_{p-1}\left[v(t_i,\overline{s}_j)\right]^c\}\ .$$

Note that, we form a set of *p*-spatially nearest voxels, $\eta_p\left[v(t_i,\overline{s}_j)\right]$ for each voxel $v(t_i,\overline{s}_j)$. Next, we define a local mesh with a star topology, where each voxel $v(t_i,\overline{s}_j)$ is placed at the center of the mesh. The surrounding nodes in the star mesh are identified by including the *p-nearest neighbors* of the center voxel. Therefore, the star mesh $\Upsilon\left(\eta_p\left[v\left(t_i,\overline{s}_j\right)\right]\right)=\left(v\left(t_i,\overline{s}_j\right)\in\eta_p,\ a_{i,j,k}\in A\right)$ consists of a center voxel and the surrounding voxels in $\eta_p$ neighborhood. Each center voxel is represented by the linear combination of its surrounding voxels,

$$v\left(t_i,\overline{s}_j\right)=\sum_{\overline{s}_k\in\eta_p} a_{i,j,k}v\left(t_i,\overline{s}_k\right)+\varepsilon_{i,j} \ , \tag{1}$$

where $a_{i,j,k}$ is the arc weights of the mesh and $\varepsilon_{i,j}$ indicates the error of voxel intensity value $v\left(t_i,\overline{s}_j\right)$ at time instant $t_i$, which will be minimized with respect to the arc weights $a_{i,j,k}$. This is achieved by minimizing the square error defined as follows,

$$\varepsilon_{i,j}{}^2=\left(v\left(t_i,\overline{s}_j\right)-\sum_{\overline{s}_k\in\eta_p\left(\overline{s}_j\right)} a_{i,j,k}v\left(t_i,\overline{s}_k\right)\right)^2 \ . \tag{2}$$

One of the standard techniques, such as Levinson-Durbin recursion (Vaidyanathan, 2008) can be employed to minimize (2) with respect to $a_{i,j,k}$. The estimated mesh arc weights $a_{i,j,k}$, $k=1,2,\ldots,p$ represent the linear relationship between the center voxel and its *p-nearest surrounding neighbors*. The estimated mesh arch weights $a_{i,j,k}$ provide us a compact representation of the relationship of each voxel $v\left(t_i,\overline{s}_k\right)$ and its surroundings $v\left(t_i,\overline{s}_j\right)$, in the *p-neighborhood* system, at a time $t_i$.

In the above formulation, each voxel $v\left(t_i,\overline{s}_j\right)$ is represented by the star mesh $\Upsilon\left(\eta_p\left[v\left(t_i,\overline{s}_j\right)\right]\right)=\left(v\left(t_i,\overline{s}_j\right)\in\eta_p,\ a_{i,j,k}\in A\right)$, which is defined over the neighborhood $\eta_p$ (see Figure 2). If we increase *p*, then the mesh size is also increased, including more neighborhood relations. For *p=0*, the mesh is reduced to a single voxel, which is identical to classical

MVPA representation of fMRI data. A critical problem is to find an “optimal” mesh size. A simple method to estimate the mesh size is to use one-leave-out cross validation applied to fMRI data. However, note that as $p$ increases, the problem becomes exponentially expensive. For practical reasons, we did not seek for an optimal p-value and suffice to test the performance of the proposed method for a mesh size of 6.

Next, we obtain an $Nxp$ dimensional Mesh Arc Descriptor (MAD) vector, $A_i = \left[\overline{a}_{i,1}\ \overline{a}_{i,2}\ \ldots\ \overline{a}_{i,N}\right]^T$ for each time, $t_i$, for $i=1,\ldots,M,$ after the optimization of (2), where, $\overline{a}_{i,j} = \left[a_{i,j,1}\ a_{i,j,2} \ldots a_{i,j,p}\right]$, for $j=1,\ldots,N$ , is the arc weights of the mesh formed around the voxel $v\left(t_i, \overline{s}_j\right)$. Recall that $p$ is the mesh size, $M$ is the number of time samples and $N$ is the number of active voxels.

In Mesh Learning, a classifier is trained by the MAD vector, $A_i$, which is extracted from the examples of the encoding stage as described in Algorithm (1). Then, the classifier is tested by the MAD vectors extracted from the examples recorded in the decoding stage. The accuracy of the classifier is measured by the ratio of the number of the correctly decoded examples to total number of examples.

**Algorithm 1 :** Extract MAD vectors**;** $A_i$

Input: Dataset : $D=\left\{v\left(t_i,\bar{s}_j\right)\right\}$, for $i=1,2,....,M$ and $j=1,2,....,N$

Order of Mesh Model : p

**Begin**

1. ***for*** i=1 to N
2. ***for*** j=1 to M
3. Compute mesh $\Upsilon\left(\eta p\left[v\left(t_i,\bar{s}_j\right)\right]\right)=\left(v\left(t_i,\bar{s}_j\right)\in\eta p, a_{i,j,k}\in A\right)$ of $v\left(t_i,\bar{s}_j\right)$ ;
4. Compute $\bar{a}_{i,j}$ optimizing (2);
5. ***endfor*** (j)
6. Construct $A_i$ using $\bar{a}_{i,j}$ ;
7. ***endfor*** (i)

**End**

Output: MAD set: $\{A_i\}_{i=1}^{N}$

In the current investigation, we evaluated the proposed model using four widely available classifiers, namely, Support Vector Machines (SVM), k-Nearest Neighbor (k-NN), Naïve Bayes and Neural Networks.

## 4. Results and Discussion

A block diagram which summarizes the proposed cognitive process classification methods is given in Figure 6. Experiment design, data acquisition and pre-processing, and anatomic feature selection methods are given in the previous sections. In this section, we focus on algorithms for feature selection, extraction and classification.

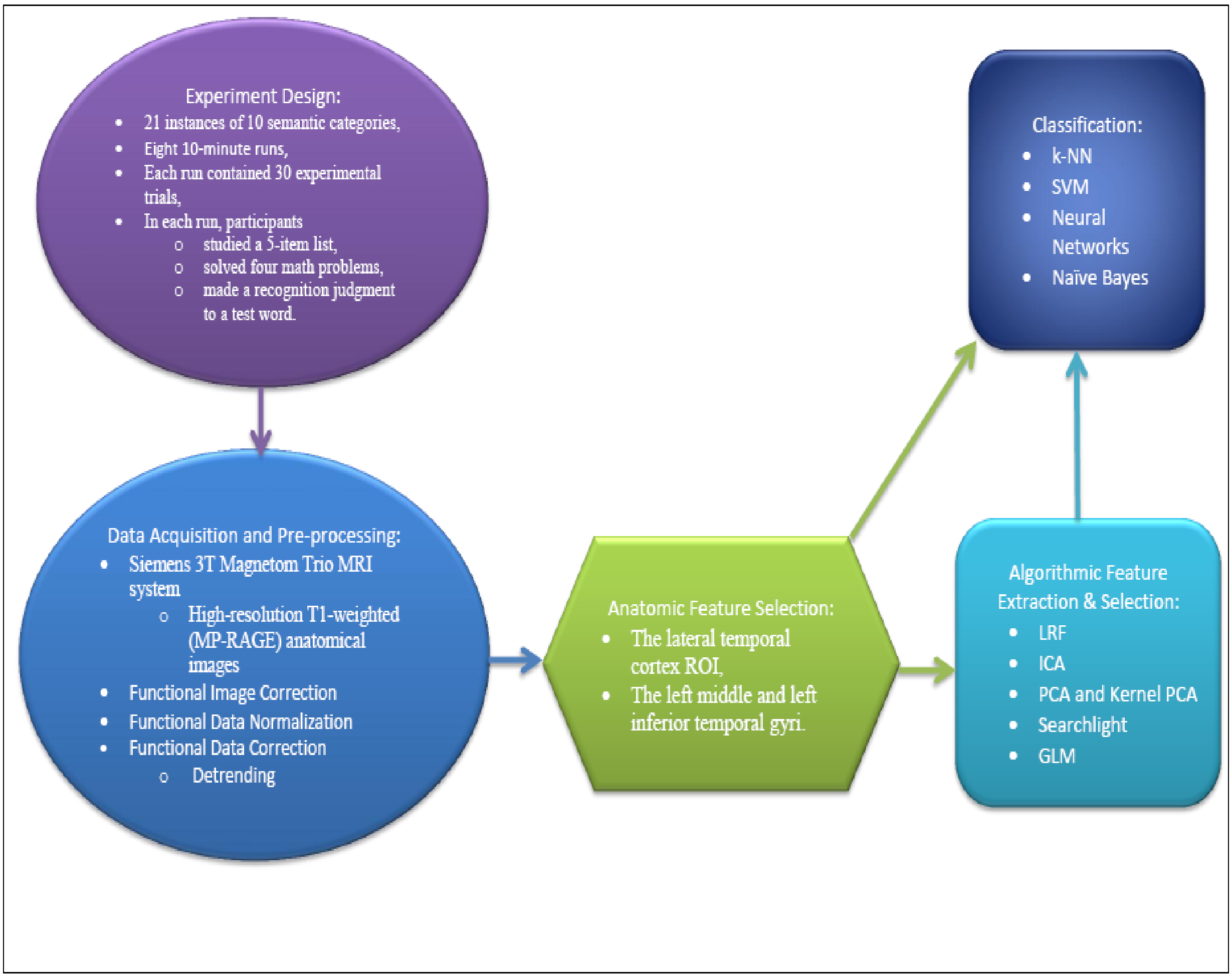

*Figure 6.* A block diagram which summarizes the proposed cognitive process classification methods.

### *4.1 Implementation of MVPA methods*

In order to evaluate the proposed Mesh Learning model, several MVPA methods were implemented. As mentioned above, our region of interested consisted of the lateral temporal cortex (LTC). In the implementation of classification algorithms, we refer to this structurally defined ROI consisting of the entire LTC as *Raw Features*. In addition, classification algorithms were also employed on the features obtained from algorithmic feature selection and extraction algorithms that are further employed on the LTC ROI.

We applied the proposed Mesh Learning method to our data for mesh size with p=6 to consider the nearest neighbors of each voxels that are spatially distributed in the 3D space of fMRI data. First, we extracted the MAD vectors to represent the information about the relationships between the spatially distributed voxels intensity measurements for each example. In order to further test the proposed Mesh Model, we also implemented three well-known feature extraction algorithms, namely, ICA, PCA and Kernel PCA. It is important to note the difference in approach across PCA, ICA and MAD vectors. In PCA, the voxel intensity measurements with maximal statistical variance are identified. That is, the features obtained from PCA provide distinct information about the statistics of the voxel intensity measurements. On the other hand, ICA maximizes the statistical independence between the voxel intensity measurements. Therefore, the features (or voxels) obtained from the ICA are statistically independent.  MAD vectors estimated in the Mesh Model describe the linear relationship between a voxel and its neighbors.

In the implementations, we used PCA-st and Kernel PCA-st implementations of Sidhu et al. (Sidhu et al., 2012) on the raw features over space and time without Fast Fourier Transform.

We used Fast ICA implementation (Hyvarinen, 1999) for ICA. In each iteration, we compute the rank of the training and test data matrices as equal to the number of training and test samples, which are $M_{tr}$ and $M_{te}$, respectively. In other words, the number of non-zero principle and independent components of PCA-st and ICA which are computed using training and test data matrices is equal to $M_{tr}$ and $M_{te}$ respectively. Therefore, the parameters of PCA-st and Fast ICA algorithms, such as the dimensions of the feature spaces to which the Raw Features are mapped, are selected using cross-validation by linearly searching the dimensions in an interval [1, $M_{tr}$]for training data, and [1, $M_{tr}$] for test data. .  The kernel type of the Kernel PCA is selected as the Gaussian Kernel, and the kernel width parameter was optimized by cross-validation. Since the number of non-zero elements of the eigenvectors of the Kernel matrix of Kernel PCA-st is computed as 900, the dimension reduction parameters of Kernel PCA-st is selected in an interval [1,900] using cross-validation.

For the implementation of GLM, separate regressors were generated for each condition and were modeled using a canonical hemodynamic response function and its temporal derivative. Data across runs were concatenated and modeled as one session with mean signal and scanner drift entered as a covariate. Using a fixed-effects model, a voxel-wise contrast identified voxels that showed more activation for the experimental conditions against the baseline at an uncorrected threshold of $p < .001$.

For the implementation of Searchlight, we employed the Searchmight (Pereira & Botvinick, 2011). In the spatial voxel selection phase of the searchlight, we selected the voxels that reside in a neighborhood of 2x2x2 of each voxel. Then, we employed a Gaussian Naïve Bayes (GNB) classifier to measure the voxel scores considering the classification accuracies. Classification accuracies of the GNB classifiers employed by each voxel were considered as the voxel scores for feature selection, i.e. the voxels with classification accuracies equal to or greater than a threshold $Acc \geq \Psi_{acc}$ were selected as the features, and the selected features were fed to the classifiers.

To evaluate the classification performance across these methods, we employed several well-known classifiers, namely, Neural Networks (NN), Gaussian Naïve Bayes, Support Vector Machines (SVM), and k-Nearest Neighbor (k-NN) methods. To be consistent with previous research, (e.g., Öztekin & Badre, 2011), we used a two-layer Back Propagation algorithm for the NN classifier, implemented in the Princeton Multi-Voxel Pattern Analysis Toolbox (http://code.google.com/p/princeton-mvpa-toolbox/). We used both Linear and RBF Kernels for the implementation of SVM (Chang and Lin, 2011). In addition, we employed Kernel density estimation method to estimate likelihood or class conditional density, in the implementation of GNB classifier (Duda et al., 2001).

In all of the implementations, classifier parameters, including the number of nearest neighbors k of k-NN, kernel parameters of RBF Kernels of SVM and kernelized GNB were selected using leave one out cross-validation on training data. In the line search optimization of the classifier parameters, k values were considered in the interval $[1, \sqrt{N}]$, where $N$ is the number of training samples, the kernel width parameter σ is searched in the interval $\log(\sigma) \in [-10, 5]$ and the SVM cost penalization parameter c is searched in the interval $\log(c) \in [-10, 5]$. The parameters of the NN were selected as the default learning rate and termination time parameters in the Princeton Multi-Voxel Pattern Analysis Toolbox.

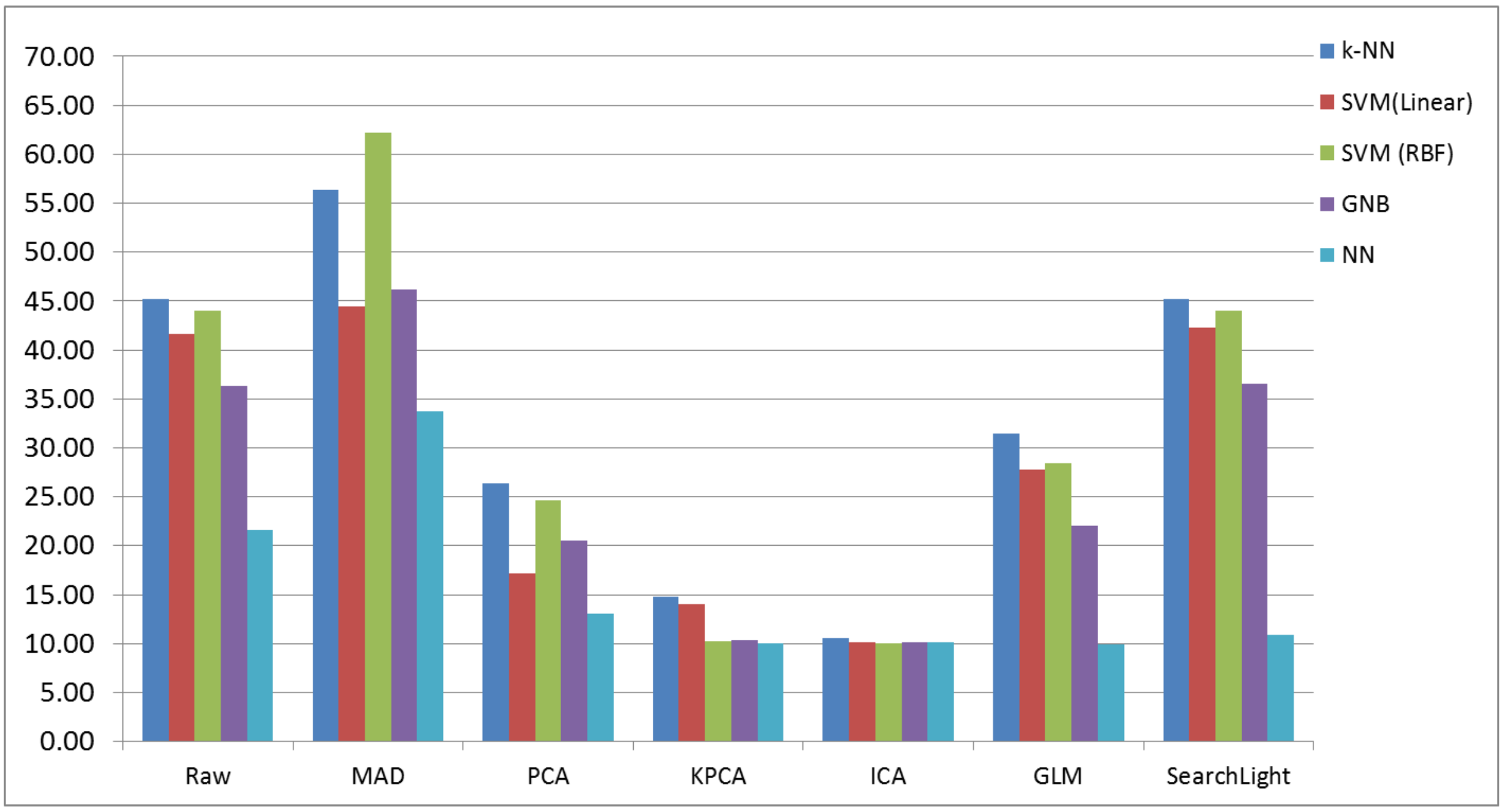

*Figure 7*. Classification accuracy in percentage of correct classification for the machine learning methods investigated across the five classifiers.

### 4.2 *Experimental Results*

A leave-one-out (N-minus-one) run-by-run cross validation provided eight iterations of classification performance for each method. Figure 3 illustrates the classification results for each method across the five classifiers.

An inspection of Figure 7 indicates that the Mesh Model and the Searchlight approaches provide relatively better classification performance compared to GLM, PCA, Kernel PCA and ICA methods. Accordingly, we limit our statistical evaluations to Searchlight and the Mesh Model. A Model [Mesh Model, Searchlight] X Classifier [k-NN, SVM Linear Kernel, SVM RBF Kernel, Naive Bayes, Neural Networks] Analysis of Variance indicated a reliable main effect of Model [$F(1,7)$ =641, p < .001], with Mesh Model yielding greater classification accuracy overall; a reliable main effect of Classifier [$F(4,7)$ =400, p < .001], and a Model X

Classifier interaction [$F(4,7) = 64.98$, $p < .001$]. Pair-wise comparisons indicated that the Mesh Learning Model had higher classification accuracy compared to Searchlight for all classifiers [$t$ = 43.18 for k-NN; $t$ = 2.73 for SVM Linear Kernel; $t$ = 21.58 for SVM RBF Kernel; $t$ = 6.25 for Naive Bayes; and $t$ = 19.44 for Neural Networks].

Notably, the classification performances for the Mesh Model also indicate that k-NN, SVM and GNB classifiers give higher performances (an average performance of 56% for k-NN, 44% for SVM Linear, 62% for SVM RBF and 46% for GNB) compared to Neural Networks (an average performance of 33%). This performance difference might result from the nature of the classification algorithms. Specifically, k-NN algorithm employs a local learning rule in which the local relational information can be successfully used for learning. Similarly, SVM methods avoid the data scarcity problems (e.g., Duda et al., 2001). Additionally, a GNB classifier, which employs kernel density estimation, provides better approximations to true class conditional densities than the non-kernelized estimates (Duda et al., 2001).

Classification performances depicted in Figure 7 also indicate that the non-linear classifiers (i.e. k-NN and SVM with RBF Kernel) provide more robust performance compared to the linear classifiers (e.g. SVM with Linear Kernel and NN). Since the variables of the optimization algorithms such as the weight vectors are linear functions of feature vectors, specifically in the optimization problem of SVM with Linear Kernels (Duda et al., 2001), the performance of these classifiers are more sensitive to the variation of the statistical properties of the feature vectors than the non-linear classifiers (Bottou & Bousquet, 2008). Accordingly, the current investigation indicates that the proposed Mesh Model approach is optimal with k-NN and SVM with RBF Kernel classifiers.

## 5. Conclusion

In this paper, we proposed a new approach, namely Mesh Learning, for pattern analysis of brain activity during cognitive processing. We tested the proposed Mesh Model's performance in classifying the semantic category of information represented in the brain during encoding and retrieval from memory. The proposed method employs local meshes, defined in a neighborhood system to represent the relationships between the voxels and their *p-nearest neighbors*. The current set of results indicates that the Mesh Learning method can improve classification performance of well-known machine learning methods, such as Neural Networks, Naïve Bayes, Support Vector Machine and k-NN classifiers. The present set of results also implicates the SVM and k-NN classifiers to yield the highest performance.

It is important to note that the proposed Mesh Learning model was tested on a data set acquired in a recognition memory study, and hence the current set of results only demonstrate its success in classification of the type of information (i.e. semantic category) represented in the brain during memory encoding and retrieval operations. Future research would prove additional insight into the feasibility of the proposed model for a wider range of cognitive processes.